\PassOptionsToPackage{table,xcdraw}{xcolor}

\documentclass[sigconf]{acmart}

\usepackage{multirow}

\definecolor{BDE6CD}{HTML}{BDE6CD}
\definecolor{E4EEBC}{HTML}{E4EEBC}
\definecolor{FFF8C5}{HTML}{FFF8C5}

\AtBeginDocument{%
  }

\acmYear{2025}
\acmConference[MM '25] {Proceedings of the 33rd ACM International Conference on Multimedia}{October 27--31, 2025}{Dublin, Ireland.}

\begin{document}

\title[SDG for Multimodal Cross-Cancer Prognosis via Dirac Rebalancer and Distribution Entanglement]{Single Domain Generalization for Multimodal Cross-Cancer Prognosis via Dirac Rebalancer and Distribution Entanglement}

\author{Jia-Xuan Jiang}
\authornote{Both authors contributed equally to this research.}
\affiliation{%
  \institution{Lanzhou University $\&$ Westlake University}
  \city{Lanzhou}
  \country{China}}
\email{jiangjx2023@lzu.edu.cn}
\orcid{0009-0001-9686-4836}

\author{Jiashuai Liu}
\authornotemark[1]
\affiliation{%
  \institution{Xi'an Jiaotong University}
  \city{Xi'an}
  \country{China}}
\email{liujs@stu.xjtu.edu.cn}
\orcid{0009-0002-5964-5658}

\author{Hongtao Wu}
\affiliation{%
  \institution{Westlake University $\&$ Chinese University of Hong Kong University}
  \country{Hang Zhou, China}
 }
\email{wuhongtao@westlake.edu.cn}
\orcid{0009-0007-4863-5119}

\author{Yifeng Wu}
\affiliation{%
  \institution{Southern University of Science and Technology}
  \state{Shenzhen}
  \country{China}
}
\email{yf.wu1@siat.ac.cn}
\orcid{0009-0005-0218-9091}

\author{Zhong Wang}
\authornote{Qi Bi, Zhong Wang and Yefeng Zheng are the corresponding authors.}
\affiliation{%
  \institution{Lanzhou University}
  \city{Lanzhou}
  \country{China}}
\email{wangzhong@lzu.edu.cn}

\author{Qi Bi}
\authornotemark[2]
\affiliation{%
  \institution{University of Amsterdam}
  \city{Amsterdam}
  \country{Netherland}}
\email{q.bi@uva.nl}

\author{Yefeng Zheng}
\authornotemark[2]
\affiliation{%
  \institution{Westlake University}
  \city{Hangzhou}
  \country{China}}
\email{zhengyefeng@westlake.edu.cn}

\renewcommand{\shortauthors}{Jia-Xuan Jiang et al.}

\begin{abstract}
 Deep learning has shown remarkable performance in integrating multimodal data for survival prediction. However, existing multimodal methods mainly focus on single cancer types and overlook the challenge of generalization across cancers. In this work, we are the first to reveal that multimodal prognosis models often generalize worse than unimodal ones in cross-cancer scenarios, despite the critical need for such robustness in clinical practice. To address this, we propose a new task: Cross-Cancer Single Domain Generalization for Multimodal Prognosis, which evaluates whether models trained on a single cancer type can generalize to unseen cancers. We identify two key challenges: degraded features from weaker modalities and ineffective multimodal integration. To tackle these, we introduce two plug-and-play modules: Sparse Dirac Information Rebalancer (SDIR) and Cancer-aware Distribution Entanglement (CADE). SDIR mitigates the dominance of strong features by applying Bernoulli-based sparsification and Dirac-inspired stabilization to enhance weaker modality signals. CADE, designed to synthesize the target domain distribution, fuses local morphological cues and global gene expression in latent space. Experiments on a four-cancer-type benchmark demonstrate superior generalization, laying the foundation for practical, robust cross-cancer multimodal prognosis. Code is available at \href{https://github.com/HopkinsKwong/MCCSDG}{here}.
\end{abstract}



\maketitle

\section{Introduction}

\begin{figure}[t]
	\centering
	\includegraphics[width=0.9\columnwidth]{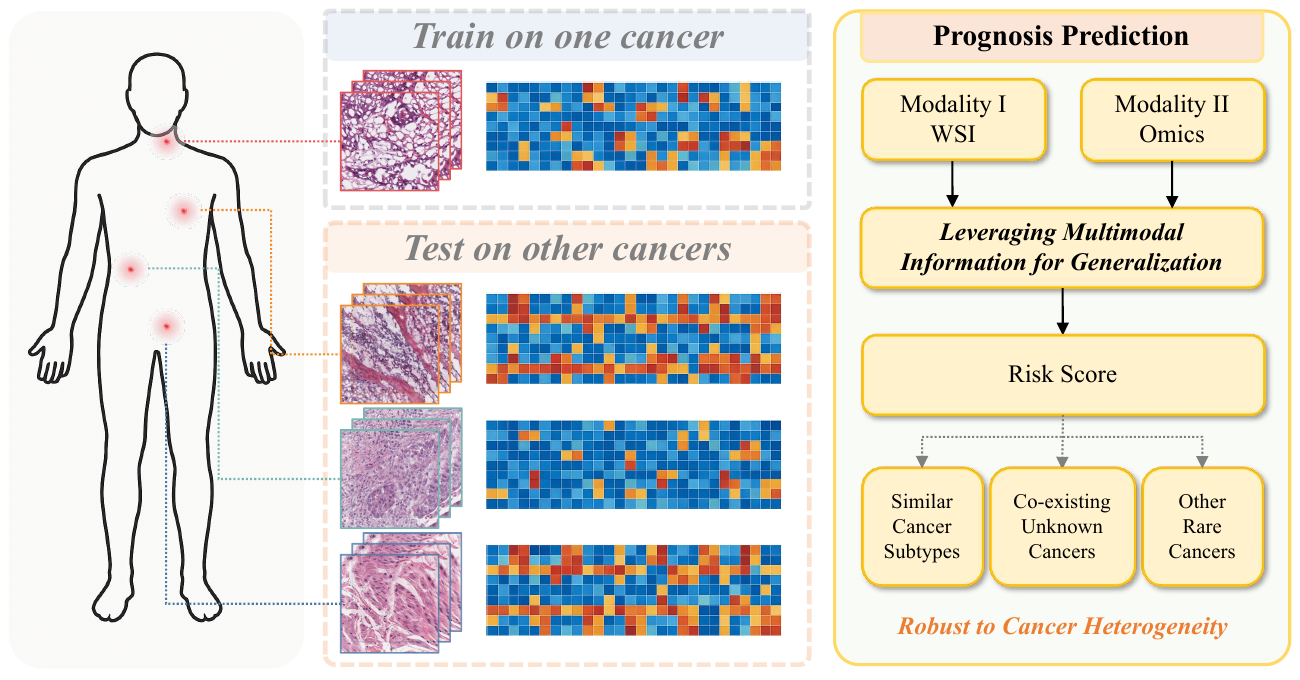}
	\caption{The left panel illustrates the task setup of single-source domain generalization for multimodal survival prediction, where a predictive model is trained on two modalities (WSI images and genomic data) from a single cancer type, and tested on other cancer types. The right panel highlights that developing multimodal survival prediction algorithms across multiple cancers can enhance the robustness and generalizability of models, especially in real-world scenarios involving rare or previously unseen cancers. However, this aspect remains largely unexplored in existing research.}
	\label{fig:task-intro}
\end{figure}

\begin{figure*}[t]
	\centering
	\includegraphics[width=0.85\textwidth]{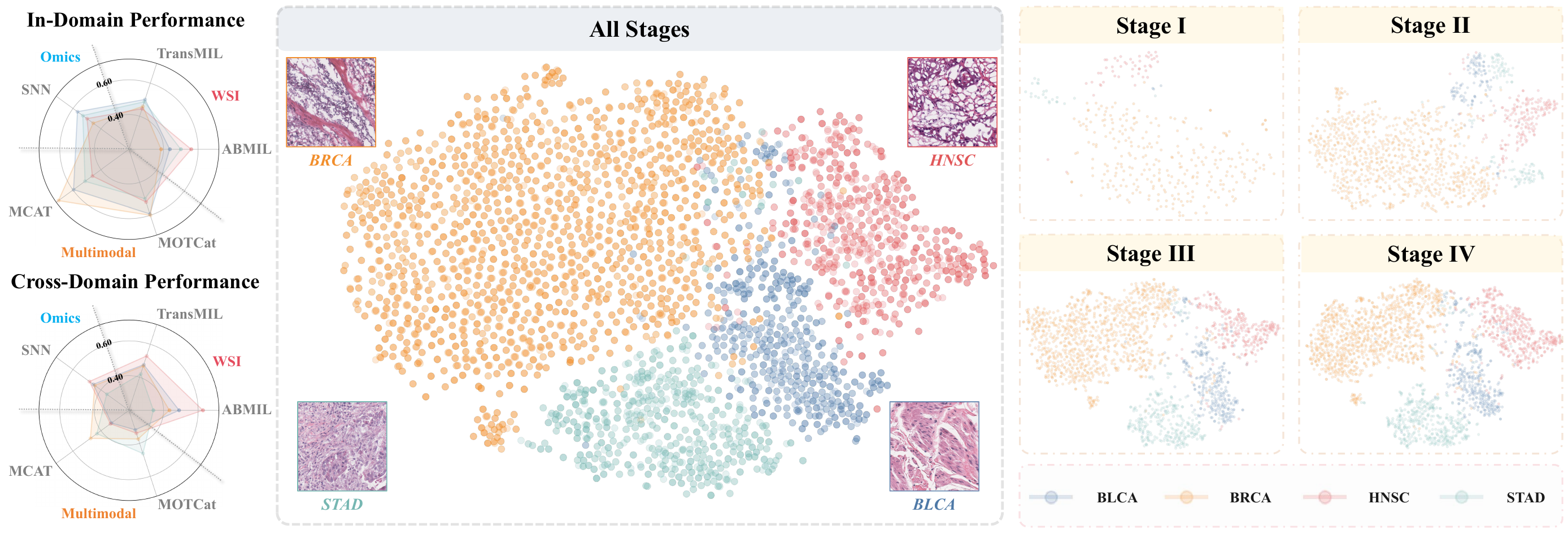}
	\caption{The left panel illustrates the performance of various prognostic methods in in-domain and cross-domain settings, revealing for the first time that although multimodal prognostic methods (i.e., MCAT and MotCat) outperform unimodal approaches (i.e., TransMIL and ABMIL with WSI, and SNN with omics), their robustness and generalizability are surprisingly inferior to those relying solely on WSI images. The question of how to enhance the generalizability of multimodal survival prognostics remains unexplored. The right panel displays the t-SNE visualization of features from WSI patches across four cancer datasets at all stages and individual stages, highlighting the significant domain shifts among different cancers, which underscores the challenge of enhancing generalizability.}
	\label{fig:intro}
\end{figure*}

Survival prediction represents a critical prognosis-related task with significant clinical implications in computational pathology \cite{prognosis1,prognosis2}. The goal is to estimate risk factors from diagnostic data to accurately rank patients based on survival time. Cancer prognosis prediction using multimodal approaches, which integrate pathological images with transcriptomic data, has shown significant potential in improving patient stratification and enabling personalized treatment \cite{multimodal_prognosis_1,li2022hfbsurv}. Despite promising advances, most existing multimodal prognostic models \cite{survpath,motcat,mcat} assume that training and testing samples originate from the same cancer type, severely limiting their applicability to more realistic clinical scenarios involving multiple cancer types. Additionally, acquiring sufficient annotated multimodal data for each cancer type remains challenging due to data scarcity, privacy concerns, and resource limitations.

Recent research on domain generalization in prognosis prediction has primarily focused on multi-source scenarios, where models trained on transcriptomic data from multiple known cancer domains (e.g., datasets from various cancer types) \cite{multiDG1} are expected to generalize to previously unseen domains. However, in many practical clinical settings, labeled data often typically exist for only one cancer type, yet clinicians still expect models to generalize effectively to other cancer (sub)types for broader applicability. For instance, a survival analysis model trained exclusively on bulk breast cancer data is frequently applied to distinct molecular subtypes, such as Luminal A and Luminal B, posing significant challenges for subtype-level generalization \cite{geneSDG}. 
Moreover, collecting all subtypes of a cancer is impractical, highlighting the pressing need of generalization across multimodal prognostic models. 

To address this challenge, we propose a novel and practically important task: \textbf{\textit{Cross-Cancer Single Domain Generalization for Multimodal Prognosis}}. 
As shown in Figure \ref{fig:task-intro}, our goal is to train a robust multimodal prognostic model using multimodal data from a single cancer type and verify whether it learns pan-cancer invariant representations, ensuring that the model remains robust when faced with unknown subtypes of the same cancer, coexisting other cancers, or rare cancers, thus meeting actual clinical needs.
The key research question to approach this task naturally arises: \textbf{\textit{how to effectively leverage pathology and genomic information to improve generalization across cancer types?}} 

Our work begins with two key observations, each revealing a distinct challenge. On one hand, as shown in Figure~\ref{fig:intro}~(left), multimodal methods unexpectedly generalize worse than single-modality methods, primarily due to a pronounced imbalance in feature quality: whole slide image (WSI) features derived from pretrained foundation models \cite{lu2024visual,chen2024towards,vorontsov2024virchow,jiang2025multi,wang2022transformer} significantly outperform gene expression features learned from scratch, adversely affecting multimodal fusion. This imbalance naturally leads to the question: \textbf{\textit{how can we effectively learn from weaker features to promote meaningful and robust multimodal integration?}} On the other hand, as illustrated in Figure\ref{fig:intro}~(right), pathological image patch features from different cancer datasets form distinct but closely adjacent clusters in the latent space, mainly because only a minority of patches contain cancer-specific signals. In contrast, gene expression profiles offer a broader, global perspective of the cancer state. Such contrasting characteristics prompt another critical question: \textbf{\textit{how can we effectively integrate local pathological features with global gene expression information to accurately generalize to unseen cancer (sub)types?}}



To address the aforementioned challenges, we propose a novel multimodal prognostic framework tailored for single-source domain generalization. First, to mitigate the impact of low-quality gene expression features in fusion, the framework balances modality contributions by reducing over-reliance on dominant signals (e.g., WSI) and enhancing weaker ones (e.g., gene expression). Second, to bridge the domain gap between the source and unseen cancer types, it synthesizes \cite{wang2025localized} latent representations by leveraging local \cite{wu2024rainmamba,wu2025samvsr} pathological cues from image patches and global signals from gene profiles, enabling better generalization across cancer domains.

Technically, our method integrates two complementary modules—Sparse Dirac Information Rebalancer (SDIR) and Cancer-aware Distribution Entanglement (CADE)—to tackle feature imbalance and modality heterogeneity. SDIR introduces structured degradation using a Bernoulli-based sparsification mask \cite{wu2023mask} to suppress dominant modalities. A Dirac-inspired nonlinear function stabilizes the degraded features, allowing the model to extract richer information from weaker modalities. CADE addresses cross-cancer generalization by constructing a kernel-smoothed statistical path between modality-specific distributions, guided by a semantic scalar. This forms a cancer-aware latent Gaussian distribution that integrates local morphology (from WSI) and global biology (from gene expression), improving adaptability to unseen cancer types. Through rigorous experimental evaluation and detailed analyses, our study demonstrates both the feasibility and clinical relevance of our approach. Concretely, our contributions can be summarized as threefold.

\begin{itemize}
    \item We propose a new task, namely, \textbf{\textit{Cross-Cancer Single Domain Generalization for Multimodal Prognosis}}, which can generalize well from only a single-source domain and is versatile to multiple cancer types. 
    \item We propose a multimodal prognostic framework tailored for the single-source generalization scenario, consisting of two key components: \textbf{\textit{Sparse Dirac Information Rebalancer}} (SDIR) and \textbf{\textit{Cancer-aware Distribution Entanglement}} (CADE). SDIR mitigates over-reliance on dominant modalities and enhances weaker signals through Bernoulli-based sparsification and a Dirac-inspired nonlinear transformation.
    \item To better exploit complementary multimodal information, we introduce CADE, which synthesizes latent target-domain distributions by integrating local pathological cues from image patches with global semantic signals from gene profiles. This structured fusion enables improved generalization across diverse cancer types.
    \item Extensive experiments validate the robustness and generalization capability of multimodal features extracted from pathological and transcriptomic data across diverse cancers.
\end{itemize}

We hope our findings could provide meaningful insights and practical guidance for developing truly generalizable multimodal prognostic models, ultimately supporting broader and more effective clinical applications.

\section{Related Work}

\subsection{Domain Generalization}

Domain generalization (DG) is a critical challenge in the real-world deployment of deep learning models \cite{MSDG3,li2025domain,wang2025illuminating,SDG1,wang2024shallow,bi2025nightadapter,bi2025learning}. Unlike domain adaptation \cite{zhu2025improving,liu2025style,yang2024genuine,jiang2025mfm}, which relies on access to target domain data for alignment, it aims to train models that can generalize well to unseen target domains without any target data available. In the medical domain, due to heterogeneity across institutions, imaging devices, and patient populations, severe domain shifts commonly exist. Most prior studies have focused on multi-source domain generalization (MSDG) \cite{MSDG1,MSDG2}, where models are trained on multiple labeled source datasets and evaluated on an unseen target domain. However, MSDG requires access to data from multiple source domains, which is often impractical in clinical practice due to high data acquisition costs and privacy concerns \cite{MSDG3}.

\subsection{Single Domain Generalization}
In contrast to MSDG, single-source domain generalization (SDG) \cite{bi2024learning,wang2025multi,SDG1,SDG3} offers a more practical setting, where models are trained using labeled data from only a single source domain and then deployed on unknown target domains. Nevertheless, differences across data centers and imaging modalities (e.g., Magnetic Resonance Imaging and Computed Tomography, abbreviated as MRI and CT, respectively) lead to distribution shifts that cause the model to overfit to the source domain, thereby degrading its performance on the target domain. Existing approaches to mitigate domain shifts generally fall into two categories: image-level \cite{SDG3} and feature-level \cite{SDG1,DG3,SDG2,wu2024semi} methods. Image-level methods aim to expand the source distribution through data augmentation to better match the target domain. In contrast, feature-level methods seek to learn domain-invariant representations via feature disentanglement, thereby improving robustness to domain-specific variations.

Although these methods have demonstrated promising results in classification and segmentation tasks, they are not specifically designed for multi-modal survival prediction, a task that presents unique challenges such as weak feature degradation and the integration of local WSI patch features, which capture fine-grained spatial information, with global gene expression features that reflect overarching biological signals. To the best of our knowledge, no existing domain generalization methods explicitly tackles the challenge of cross-cancer multi-modal survival prediction under single-source constraints. Our work fills this critical research gap.

\subsection{Multi-Modal Prognosis Prediction}

Accurate cancer survival prediction remains a critical challenge due to the complex and heterogeneous nature of tumor biology. Histopathology images and transcriptomic profiles capture complementary phenotypic and molecular characteristics, and have thus become key modalities for predictive modeling. For single-modality approaches, multiple instance learning (MIL) methods applied to pathology images \cite{gao2024accurate,lu2021data,lu2021data,transmil2021} enable effective extraction of histological features from whole slide images, while transcriptomic-based models \cite{knottenbelt2024coxkan,hao2018pasnet} leverage deep neural networks to uncover prognostic signals from RNA-seq data. To harness the strengths of both modalities, recent studies \cite{wang2021gpdbn,li2022hfbsurv} have proposed multimodal survival prediction frameworks. Late fusion strategies, which combine predictions from separate unimodal models, offer an intuitive and straightforward solution for multimodal integration, but often overlook cross-modal dependencies. In contrast, early fusion methods \cite{zhang2024prototypical,zhou2023cross,mcat,motcat,survpath} introduce more sophisticated integration schemes based on cross-attention mechanisms. Specifically, Chen et al. \cite{mcat} enable fine-grained feature fusion via genomic-guided cross-attention; Xu et al. \cite{motcat} applie optimal transport to align heterogeneous representations; and Jaume et al. \cite{survpath} incorporate biological prior knowledge by tokenizing gene expression data along pathway structures for meaningful fusion with histological features.

Although these multimodal approaches achieve strong performance on individual datasets, we observe a notable drop in generalization under domain shifts. In comparison, unimodal models exhibit more consistent behavior across datasets. Motivated by this, we propose a novel framework designed for single-domain generalization, aiming to preserve the generalizability of unimodal models while effectively capturing cross-modal complementarities.

\section{Methodology}

\begin{figure*}[t]
	\centering
	\includegraphics[width=0.85\textwidth]{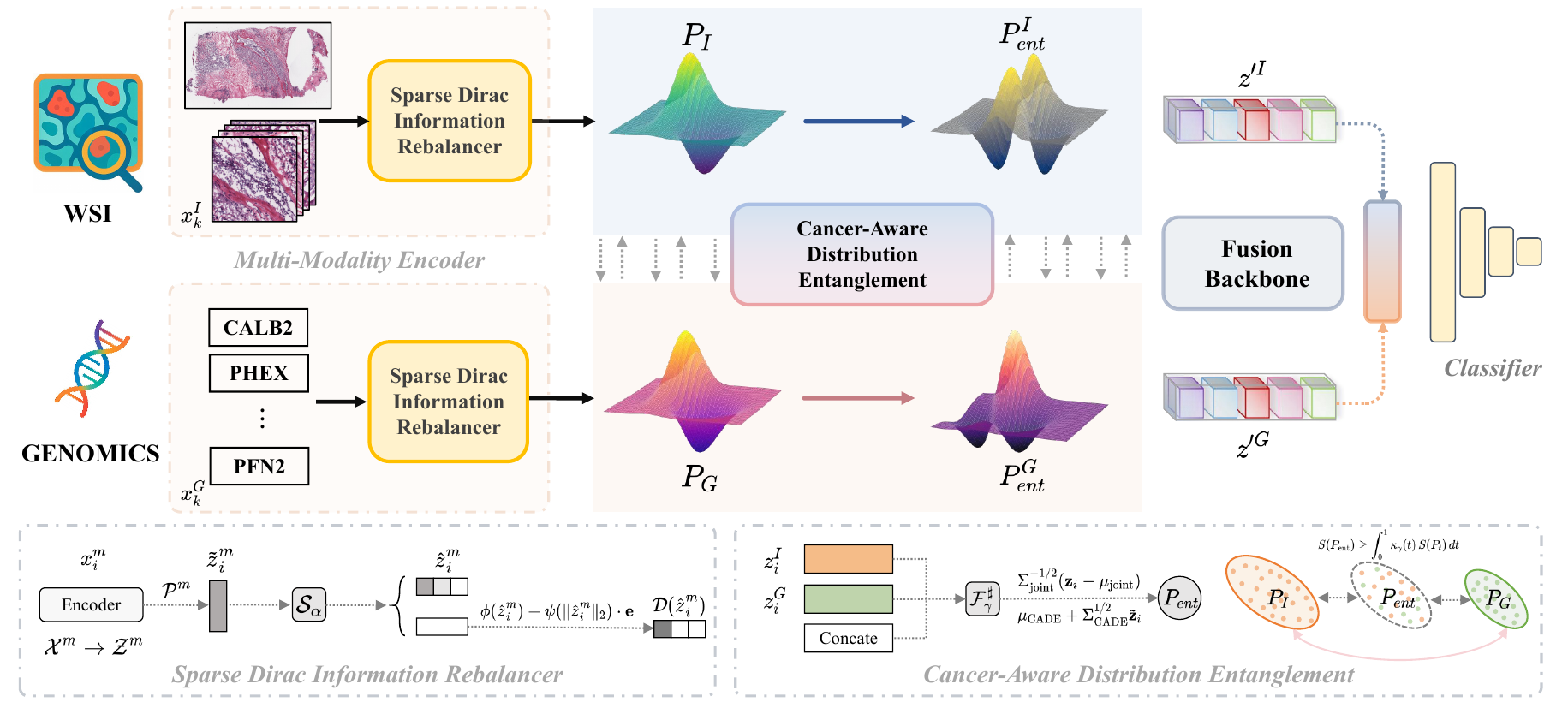}
	\caption{The proposed method aims to enhance the model's robustness in addressing weak feature degradation by achieving more reliable fusion. It leverages the local information from pathological images and the global information from gene expression to model the target domain distribution in the latent space, thereby improving the model's generalization capability. The workflow of the method includes: Sparse Dirac Information Rebalancer, Cancer-Aware Distribution Entanglement, and a multimodal feature fusion backbone network for classification.}
	\label{fig:model}
\end{figure*}

\subsection{Preliminaries \& Framework Overview}
In practical clinical scenarios, it is often impractical to obtain comprehensive subtype annotations for all cancer types due to limited data availability and resource constraints. Even within a single cancer type, such as lung cancer, substantial molecular heterogeneity exists—for example, adenocarcinoma and squamous cell carcinoma exhibit distinct morphological and genomic profiles \cite{de2018heterogeneity}. This intra-type diversity poses significant challenges for building generalized prognostic models, as a single model may fail to capture subtype-specific patterns. If the model can enhance its generalization ability by utilizing multi-modal information to model the latent distribution of cancer target domains, this suggests that it is capable of achieving reliable generalization across different cancer types. Consequently, it should also be able to adapt to the variations among subtypes within the same cancer type — indicating that generalization across cancer types also reflects the model’s robustness to intra-tumor heterogeneity.

Based on this insight, we formalize the \textbf{Cross-Cancer Single Domain Generalization for Multimodal Prognosis} task. Suppose we have WSI patch data \( x^I \), gene expression data \( x^G \), and corresponding survival outcome labels \( y \), collected from \( K \) different cancer domains. We define multiple datasets $\mathcal{D}_1, \mathcal{D}_2, \dots, \mathcal{D}_K$, where each dataset $\mathcal{D}_k$ consists of $N_k$ samples in the form of $\left(x_n^{I,(k)}, x_n^{G,(k)}, y_n^{(k)}\right)$, representing input $x_n^{I,(k)}$ and $x_n^{G,(k)}$, along with the corresponding target labels $y_n^{(k)}$. In the context of multi-modal prognosis prediction under single-source domain generalization, our goal is to train a prediction model \( F_\theta: (x^I, x^G) \rightarrow y \) using only the source domain \( \mathcal{D}_1 \), while ensuring strong generalization performance on unseen target cancer domains \( \mathcal{D}_2, \ldots, \mathcal{D}_K \). Analogous to prior cross-modality segmentation tasks where models are trained on MRI and tested on CT \cite{SDG3}, in our study, each cancer type dataset is treated as a separate domain \( \mathcal{D}_k \).

Our framework comprises two key components: the Sparse Dirac Information Rebalancer, which mitigates over-reliance on dominant modalities through Bernoulli-based sparsification and Dirac-inspired nonlinear transformations to enhance weak features; and the Cancer-aware Distribution Entanglement, which integrates local pathological cues from image patches with global genomic signals from gene profiles to generate the underlying distribution of potential target domains. This framework effectively addresses two critical challenges in cross-cancer single-domain generalization for multimodal prognosis: (1) the degradation of weak features, and (2) the enhancement of generalization by leveraging multimodal information, as detailed below.

\subsection{Sparse Dirac Information Rebalancer}

In multimodal cancer prognosis, effectively integrating information from diverse modalities remains challenging due to significant feature quality imbalance. Features extracted from whole-slide images via pretrained foundation models typically exhibit high robustness and consistency, whereas gene expression features—learned from scratch using a simple MLP—often suffer from instability and susceptibility to noise. This imbalance causes models to disproportionately depend on robust modalities, marginalizing weaker but potentially complementary features and thus undermining overall predictive effectiveness.

To address this challenge, we propose the Sparse Dirac Information Rebalancer (SDIR) module, specifically designed to rebalance modality contributions through structured sparsification and enhancement mechanisms. Instead of solely degrading strong modality signals, SDIR emphasizes explicit correction and amplification of weaker modality features, thereby facilitating balanced and robust multimodal integration.

Given multimodal data samples from modality $m \in \{I, G\}$ (where $I$ represents WSI and $G$ denotes gene expression), we first encode each input through modality-specific encoders:
\begin{equation}
E^m: \mathcal{X}^m \rightarrow \mathcal{Z}^m, \quad z_i^m = E^m(x_i^m).
\end{equation}

Encoded representations are then projected into a unified latent space to facilitate direct modality comparison and integration:
\begin{equation}
\mathcal{P}^m: \mathcal{Z}^m \rightarrow \mathcal{Z}, \quad \tilde{z}_i^m = \mathcal{P}^m(z_i^m).
\end{equation}

To rebalance modality contributions and prevent over-reliance on stronger modalities, we introduce a sparsification operator selectively applied to robust modality representations (e.g., WSI). This operator employs a Bernoulli-based mask to introduce structured sparsity:
\begin{equation}
\begin{aligned}
& \hat{z}_i^m = \mathcal{S}_\alpha(\tilde{z}_i^m) = \mathbf{D}_\alpha \cdot \tilde{z}_i^m, \quad  \\
&\mathbf{D}_\alpha = \text{diag}(\mathbf{\nu}), \quad \\
&\mathbf{\nu} \sim \text{Bernoulli}(1 - \alpha).
\end{aligned}
\end{equation}
Here, the sparsity parameter $\alpha \in [0,1)$ controls the degree of rebalancing. By systematically reducing reliance on strong signals, the model is incentivized to exploit and enhance discriminative features from weaker modalities.

To actively enhance weaker modality signals—particularly under conditions of potential feature collapse—we employ a Dirac-inspired nonlinear response function. This strategy utilizes the Dirac delta distribution concept, modeling signal concentration or collapse:
\begin{equation}
\mathcal{D}(\hat{z}_i^m) = \phi(\hat{z}_i^m) + \psi(\|\hat{z}_i^m\|_2) \cdot \mathbf{e}.
\end{equation}
Here, $\phi(\cdot)$ denotes a lightweight linear transformation, while $\psi(r) = \exp(-r)$ serves as a monotonic decay function, where $r$ is actually the L2 norm of $\hat{z}_i^m$. The template vector $\mathbf{e} \in \mathbb{R}^d$ provides a stable anchor, preserving meaningful representations even under severe degradation:
\begin{equation}
\mathcal{D}(\hat{z}_i^m) \rightarrow \mathbf{e}, \quad \text{as } \|\hat{z}_i^m\|_2 \rightarrow 0.
\end{equation}
Conversely, when sufficient signal strength is present, the function approximates identity, maintaining informational integrity:
\begin{equation}
\mathcal{D}(\hat{z}_i^m) \approx \phi(\hat{z}_i^m).
\end{equation}

This mechanism explicitly strengthens weaker features, mitigating information degradation and ensuring robust latent representations across varying modality quality conditions.

The complete definition of the SDIR module is:
\begin{equation}
\text{SDIR}_\alpha(x_i^m) := \mathcal{D} \circ \mathcal{S}_\alpha \circ \mathcal{P}^m \circ E^m(x_i^m).
\end{equation}

In summary, by strategically rebalancing modality influences and explicitly enhancing weaker signals, SDIR effectively addresses feature quality imbalances, promoting meaningful, robust, and balanced multimodal integration to improve cancer prognosis accuracy.

\begin{table*}[htbp]
\caption{Results of our proposed method and baselines in predicting cancer-specific patient survival in single-source domain generalization, measured with C-index. All omics and multimodal baselines were trained with the Reactome and Hallmark pathway sets. Top three results are highlighted as \colorbox{BDE6CD}{\textbf{best}}, \colorbox{E4EEBC}{second} and \colorbox{FFF8C5}{third}, respectively.}
\label{tab:main}
\resizebox{0.85\textwidth}{!}{%
\begin{tabular}{c|c|c|cccc|c}
\hline \hline
Modality                      & Model/Study & Publications & BLCA$\to$Target I                    & BRCA$\to$Target II                     & HNSC$\to$Target III                  & STAD$\to$Target IV                   & Average                                 \\ \hline
Multimodal                    & Upper Bound & -            & 0.6866$\pm$0.0353                                  & 0.6081$\pm$0.0349                                  & 0.5584$\pm$0.0792                                  & 0.5145$\pm$0.0912                                  & 0.5919                                  \\ \hline
                              & ABMIL \cite{ABMIL}       & ICML2018     & 0.5450$\pm$0.0708                                  & 0.5169$\pm$0.0237                                  & \cellcolor[HTML]{BDE6CD}\textbf{0.6137$\pm$0.0249} & 0.4707$\pm$0.0714                                  & 0.5366                                  \\
\multirow{-2}{*}{WSI}         & TransMIL \cite{transmil2021}    & NIPS2021     & 0.5384$\pm$0.0393                                  & 0.5345$\pm$0.0278                                  & \cellcolor[HTML]{FFF8C5}0.5646$\pm$0.0340          & 0.5104$\pm$0.0370                                  & 0.5370                                  \\ \hline
                              & MLP         & -            & \cellcolor[HTML]{BDE6CD}\textbf{0.6145$\pm$0.0562} & \cellcolor[HTML]{FFF8C5}0.5453$\pm$0.0321          & 0.5550$\pm$0.0199                                  & 0.4809$\pm$0.0237                                  & \cellcolor[HTML]{E4EEBC}0.5489          \\
\multirow{-2}{*}{Omics}       & SNN \cite{SNN}         & NIPS2017     & 0.5274$\pm$0.0109                                  & 0.5219$\pm$0.0142                                  & 0.5411$\pm$0.0232                                  & 0.4795$\pm$0.0364                                  & 0.5175                                  \\ \hline
                              & MCAT \cite{mcat}       & ICCV2021     & 0.4625$\pm$0.0277                                  & 0.5375$\pm$0.0356                                  & 0.4662$\pm$0.0260                                  & 0.5145$\pm$0.0014                                  & 0.4952                                  \\
                              & MOTCat \cite{motcat}     & ICCV2023     & 0.4592$\pm$0.0087                                  & 0.4869$\pm$0.0274                                  & 0.4693$\pm$0.0039                                  & 0.5309$\pm$0.0035                                  & 0.4866                                  \\
                              & SURVPATH \cite{survpath}    & CVPR2024     & 0.5255$\pm$0.0157                                  & 0.5089$\pm$0.0413                                  & 0.5385$\pm$0.0193                                  & 0.4970$\pm$0.0967                                  & 0.5175                                  \\ \cline{2-8} 
                              & Dropout \cite{Dropout}    & JMLR2014     & 0.5205$\pm$0.0333                                  & 0.5289$\pm$0.0295                                  & 0.5209$\pm$0.0390                                  & 0.4848$\pm$0.0398                                  & 0.5138                                  \\
                              & IN \cite{IN}         & arXiv2016    & 0.5140$\pm$0.0365                                  & 0.5041$\pm$0.0742                                  & 0.5366$\pm$0.0382                                  & 0.5344$\pm$0.0351                                  & 0.5222                                  \\
                              & DeepCoral \cite{DeepCoral}   & ECCV2016     & 0.5339$\pm$0.0679                                  & 0.5335$\pm$0.0376                                  & 0.5374$\pm$0.0572                                  & 0.5290$\pm$0.0804                                  & 0.5334                                  \\
                              & WT \cite{WT}         & NIPS2017     & 0.5327$\pm$0.0052                                  & \cellcolor[HTML]{E4EEBC}0.5533$\pm$0.0357          & 0.5169$\pm$0.0551                                  & \cellcolor[HTML]{FFF8C5}0.5374$\pm$0.0300          & 0.5351                                  \\
                              & MixStyle \cite{mixstyle}   & ICLR2021     & 0.5107$\pm$0.0246                                  & 0.5147$\pm$0.0600                                    & 0.5303$\pm$0.0687                                  & \cellcolor[HTML]{BDE6CD}\textbf{0.5644$\pm$0.0325} & 0.5300                                  \\
                              & BNE \cite{BNE}         & PR2023       & 0.5296$\pm$0.0107                                  & 0.5450$\pm$0.0450                                   & 0.5257$\pm$0.0195                                  & 0.5297$\pm$0.0270                                   & 0.5325                                  \\
                              & DFQ \cite{DFQ}        & AAAI2024     & \cellcolor[HTML]{FFF8C5}0.5457$\pm$0.0099          & 0.5162$\pm$0.0102                                  & 0.5586$\pm$0.0161                                  & 0.5272$\pm$0.0268                                  & \cellcolor[HTML]{FFF8C5}0.5369          \\
\multirow{-11}{*}{Multimodal} & ours        & -            & \cellcolor[HTML]{E4EEBC}0.5648$\pm$0.0477          & \cellcolor[HTML]{BDE6CD}\textbf{0.5576$\pm$0.0181} & \cellcolor[HTML]{E4EEBC}0.5762$\pm$0.0481          & \cellcolor[HTML]{E4EEBC}0.5514$\pm$0.0345          & \cellcolor[HTML]{BDE6CD}\textbf{0.5625} \\ \hline \hline
\end{tabular}%
}
\end{table*}

\subsection{Cancer-Aware Distribution Entanglement}

Another key challenge in single-source domain generalization for multimodal  prognosis lies in effectively integrating \textit{local} and \textit{global} information across heterogeneous modalities. Whole-slide image (WSI) patches encode localized morphological features, while gene expression profiles capture global signals reflecting the holistic biological state of the tumor. However, the majority of WSI patches correspond to non-cancerous regions, offering sparse or misleading information when considered in isolation.

To address this, we hypothesize that the global semantic context embedded in gene expression can serve as an informative prior to guide the structural organization of local WSI features in the latent space. Instead of directly aligning these heterogeneous modalities, we seek to construct a latent distribution that selectively absorbs cancer-relevant local details from WSI patches while remaining grounded in gene-level global semantics. We implement this idea via Cancer-aware Distribution Entanglement (CADE), a latent-space statistical modeling framework that composes a novel latent distribution using a modality-aware fusion operator. This operator aggregates modality-specific statistics in a smooth and geometry-aware fashion, resulting in a cancer-aware synthetic domain distribution suitable for generalization.

Let \( \{\mathbf{z}_i^I\} \sim P_I \) and \( \{\mathbf{z}_i^G\} \sim P_G \) denote latent features from WSI and gene modalities, respectively, with corresponding empirical statistics \( \theta_I = (\mu_I, \Sigma_I) \) and \( \theta_G = (\mu_G, \Sigma_G) \). We define a parameterized statistical composition operator \( \mathcal{F}_\gamma \) controlled by a semantic guidance scalar \( \gamma \in [0, 1] \), which softly blends these statistics via kernel-weighted integration along a modality-conditioned semantic path:

\begin{equation}
\begin{aligned}
& \mu_{\text{CADE}} = \mathcal{F}_\gamma^\mu(\mu_G, \mu_I) = \int_0^1 \mu(t) \kappa_\gamma(t) dt, \quad  \\
& \Sigma_{\text{CADE}} = \mathcal{F}_\gamma^\Sigma(\Sigma_G, \Sigma_I) = \int_0^1 \Sigma(t) \kappa_\gamma(t) dt,
\end{aligned}
\end{equation}
where $\mu(t) = (1 - t)\mu_G + t\mu_I$, $\Sigma(t) = (1 - t)\Sigma_G + t\Sigma_I$, and \( \kappa_\gamma(t) \sim \text{Beta}(\gamma, \gamma) \) is a symmetric smoothing kernel centered at \(\gamma\), normalized as \( \int_0^1 \kappa_\gamma(t) dt = 1 \). To improve numerical stability and enable fair integration across modalities, we normalize the concatenated latent features using batch-wise statistics. Given \( \mathbf{z}_i = [\mathbf{z}_i^I; \mathbf{z}_i^G] \), we compute:

\begin{equation}
\tilde{\mathbf{z}}_i = \Sigma_{\text{joint}}^{-1/2} (\mathbf{z}_i - \mu_{\text{joint}}),
\end{equation}
where \( \mu_{\text{joint}} \) and \( \Sigma_{\text{joint}} \) are the batch-wise mean and diagonal covariance of the joint latent representation. This normalization decouples modality-specific scales and ensures that statistical composition operates on standardized features. The final entangled representation is then given by:

\begin{equation}
\mathbf{z}_i' = \mu_{\text{CADE}} + \Sigma_{\text{CADE}}^{1/2} \tilde{\mathbf{z}}_i.
\end{equation}

This transformation defines a synthetic latent distribution \( P_{\text{CADE}} \in \mathcal{P}(\mathbb{R}^d) \), which we interpret as a semantic-path-aligned distributional composition:
\begin{equation}
P_{\text{ent}} := \mathcal{F}_\gamma^\sharp(P_G, P_I).
\end{equation}
This formulation bridges the global cancer-aware context of gene expression with the spatially localized yet noisy WSI features, producing a structured latent domain that implicitly spans regions not covered by the original data. To ensure that the composed distribution \( P_{\text{ent}} \) is valid, we invoke the \textit{Cramér–Wold theorem}:

\begin{theorem}[Cramér–Wold]
Let \( \{X_n\} \) be a sequence of \( d \)-dimensional random vectors, and \( X \in \mathbb{R}^d \) a target vector. Then \( X_n \xrightarrow{d} X \) (i.e., convergence in distribution) holds if and only if for every direction \( \mathbf{a} \in \mathbb{R}^d \), the projections \( \mathbf{a}^\top X_n \xrightarrow{d} \mathbf{a}^\top X \).
\end{theorem}

In our context, the smooth statistical path \((\mu(t), \Sigma(t))\) ensures convergence of all one-dimensional projections under kernel integration, thereby guaranteeing that \( P_{\text{ent}} \) defines a valid multivariate Gaussian distribution. Moreover, due to the concavity of differential entropy under integration over valid Gaussian statistics, the following inequality holds:
\begin{equation}
S(P_{\text{ent}}) \geq \int_0^1 \kappa_\gamma(t) \, S(P_t) \, dt.
\end{equation}
where \( S(\cdot) \) denotes differential entropy. This implies that CADE expands latent space entropy in expectation, leading to improved representational diversity and reduced overfitting to modality-specific priors.

CADE introduces a principled, kernel-driven statistical fusion strategy that leverages the semantic strength of gene expression to reorganize local image features into a coherent, cancer-aware latent domain. This composed latent distribution acts as a soft surrogate for target cancer domains, enhancing cross-type generalization and supporting robust multimodal inference.

\subsection{Implementation Details \& Optimization}
Biological pathways regulate a wide range of cellular processes that shape tumor morphology, including proliferation, apoptosis, and stromal remodeling \cite{qu2021genetic}. These molecular processes give rise to distinct histomorphological phenotypes observable in whole-slide images. To capture such cross-modal associations, we leverage a cross-attention mechanism between pathway-level transcriptomic features and image regions, similar to the strategy used in SurvPath \cite{survpath}. This enables the model to learn consistent prognostic signals across modalities, which are subsequently integrated through a unified survival prediction head.

To jointly optimize survival prediction, modality perturbation, and cross-modal compactness, we define the final objective function as:

\begin{equation}
\begin{aligned}
\mathcal{L}_{\text{total}}  = &
\mathbb{E} \left[ \ell(f(E(\mathbf{x})), y) \right] + 
\mathbb{E} \left[ \ell(f(\text{SDIR}_\alpha(\mathbf{x})), y) \right]  \\ & + 
\text{KL}(P_{\text{model}} \parallel P_{\text{ent}}).
\end{aligned}
\end{equation}

Here, \(\alpha\) is a sparsity parameter in SDIR, and \(\gamma\) is a semantic guidance scalar in CADE, both of which control the contribution of their respective regularization terms. The Kullback-Leibler divergence $\text{KL}(P_{\text{model}} \parallel P_{\text{CADE}})$ serves as a regularization to encourage the learned distribution.

\section{Experiments}
\subsection{Datasets \& Evaluation Protocols}

We conduct experiments on four publicly available TCGA cancer datasets. The Breast Invasive Carcinoma (BRCA) \cite{cancer2012comprehensive} dataset consists of 880 slides, the Bladder Urothelial Carcinoma (BLCA) \cite{robertson2017comprehensive} dataset contains 346 slides, the Stomach Adenocarcinoma (STAD) \cite{cancer2014comprehensive} dataset includes 322 slides, and the Head and Neck Squamous Cell Carcinoma (HNSC) \cite{cancer2015comprehensive} dataset comprises 401 slides. All histopathology images are divided into patches with $512 \times 512$ pixels at $20\times$ magnification, and features are extracted using the pretrained Conch v1.5 \cite{ding2024multimodal}. Following the methodology outlined in SurvPath \cite{survpath}, we perform gene feature selection and identify 4,999 unique genes. These genes are subsequently categorized into 331 pathways, including 281 Reactome pathways \cite{fabregat2018reactome} derived from 1,577 genes and 50 Hallmark pathways \cite{subramanian2005gene,liberzon2015molecular} derived from 4,241 genes. The performance of the model is evaluated using the concordance index (C-index) as the primary metric.

The model is trained for 100 epochs on the multimodal survival prediction task to ensure convergence. In alignment with \cite{SDG3,SDG1}, one dataset is used as the source domain, while the remaining datasets are treated as target domains. As shown in Table~\ref{tab:main}, Target I refers to the combination of BRCA, HNSC, and STAD (excluding BLCA). All experiments in our study adopt this setup. To increase data diversity during training, 4,096 patches are randomly sampled from the WSIs. During testing, all available patches are used to generate the final prediction.

\subsection{Comparison with State-of-the-art}
\subsubsection{Baselines}

We categorize the baseline methods into four groups: (1) unimodal histopathological methods, (2) unimodal transcriptomic methods, (3) multimodal methods, and (4) generalization-focused methods.

\textbf{Histopathological Baselines}: The compared approaches include:~~(a) ABMIL \cite{ABMIL}, which utilizes gated attention pooling; (b)Tra-nsMIL \cite{transmil2021}, which approximates patch-level self-attention via the Nystrom method \cite{xiong2021nystromformer}.

\textbf{Transcriptomic Baselines}: All baseline methods in this category use identical input features, which are derived by aggregating transcriptomic data from Reactome and Hallmarks pathways. (a) MLP \cite{SNN}, a four-layer multilayer perceptron; (b) SNN \cite{SNN}, which extends MLP with additional Alpha Dropout layers.

\textbf{Multimodal Baselines}: (a) MCAT \cite{mcat}, which employs genomics-guided cross-attention followed by modality-specific self-attention modules; (b) MOTCat \cite{motcat}, which applies optimal transport (OT) to align patch-level and genomic-level features.

\textbf{Generalization Baselines}: Based on state-of-the-art multimodal methods, we evaluate a variety of generalization techniques, including Dropout \cite{Dropout}, IN \cite{IN}, DeepCoral \cite{DeepCoral}, WT \cite{WT}, MixStyle \cite{mixstyle}, BNE \cite{BNE}, DFQ \cite{DFQ} and our proposed method.

\subsubsection{Generalization analysis of different unimodal methods}
As shown in Table~\ref{tab:main}, we compared the generalization performance of unimodal methods based on whole slide image and omics. Among the WSI-based models, TransMIL achieved the best average performance (0.5370), while the omics-based MLP baseline attained the highest generalization performance among all unimodal methods (0.5489). In contrast, existing state-of-the-art multimodal methods underperformed compared with unimodal approaches, which may be attributed to the weaker discriminative power of genomic features compared to WSI patch features. This imbalance can lead to feature degradation, ultimately reducing the robustness of multimodal prognostic models. Addressing this issue is one of the primary goals of the proposed method.
\subsubsection{Generalization analysis of multimodal methods}
As shown in Table~\ref{tab:main}, most multimodal methods underperformed compared to unimodal models, highlighting the limited generalization capability of existing multimodal prognostic approaches. Multimodal methods such as MCAT, MOTCat, and SURVPATH struggled with domain shift and exhibited constrained performance, with average C-index values ranging from 0.4866 to 0.5175. Although domain generalization techniques such as MixStyle and DFQ improved model robustness to some extent, they still failed to consistently surpass the best-performing unimodal methods. In contrast, our proposed method achieved an average C-index of 0.5625, outperforming all baseline models and significantly exceeding the performance of the best unimodal approaches. Specifically, our model ranked in the top two across all four target domains. This consistent cross-domain advantage demonstrates the strong generalization capability of our method under the single-source domain generalization setting across cancer types.

\subsection{Ablation Study}
\begin{figure*}[t]
	\centering
	\includegraphics[width=0.85\textwidth]{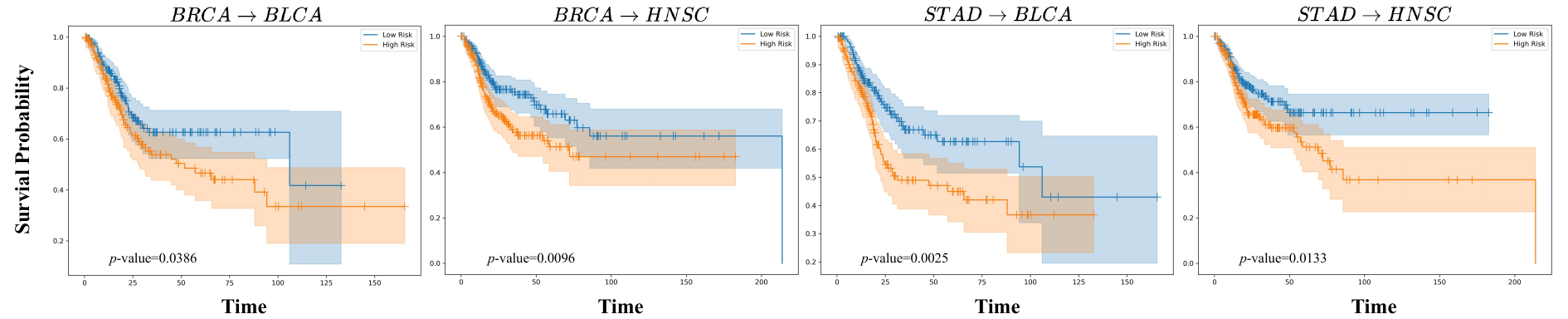}
	\caption{We utilize Kaplan-Meier analysis to evaluate the prediction performance for four cross-cancer generalization tasks. According to the predicted risk scores, all patients are stratified into a low-risk group (blue) and a high-risk group (red). }
	\label{fig:KM}
\end{figure*}

\subsubsection{Effectiveness analysis}

\begin{table}[]
\caption{Ablation study on each component. SDIR: Sparse Dirac Information Rebalancer; CADE: Cancer-aware Distribution Entanglement. Experiments are conducted on each single source domain and tested on other target domains. The evaluation metric is the C-index.}
\label{tab:ablation}
\resizebox{0.9\columnwidth}{!}{%
\begin{tabular}{ccc|cccc|c}
\hline \hline
Backbone     & SDIR          & CADE          & BLCA                                    & BRCA                                      & HNSC                                    & STAD                                    & Average                                 \\ \hline
$\checkmark$ &              &              & 0.5255                                  & 0.5089                                  & \cellcolor[HTML]{FFF8C5}0.5385          & 0.4970                                  & 0.5175                                  \\
$\checkmark$ & $\checkmark$ &              & \cellcolor[HTML]{E4EEBC}0.5456          & \cellcolor[HTML]{BDE6CD}\textbf{0.5724} & \cellcolor[HTML]{E4EEBC}0.5429          & \cellcolor[HTML]{FFF8C5}0.5307          & \cellcolor[HTML]{E4EEBC}0.5479          \\
$\checkmark$ &              & $\checkmark$ & \cellcolor[HTML]{FFF8C5}0.5405          & \cellcolor[HTML]{E4EEBC}0.5584          & 0.5056                                  & \cellcolor[HTML]{BDE6CD}\textbf{0.5567} & \cellcolor[HTML]{FFF8C5}0.5403          \\
$\checkmark$ & $\checkmark$ & $\checkmark$ & \cellcolor[HTML]{BDE6CD}\textbf{0.5648} & \cellcolor[HTML]{FFF8C5}0.5576          & \cellcolor[HTML]{BDE6CD}\textbf{0.5762} & \cellcolor[HTML]{E4EEBC}0.5514          & \cellcolor[HTML]{BDE6CD}\textbf{0.5625} \\ \hline \hline
\end{tabular}
}
\end{table}

 As summarized in Table~\ref{tab:ablation}, models are trained on a single source and tested on unseen target domains. Starting from the baseline (average C-index: 0.5175), adding SDIR leads to a 5.87\% relative improvement (0.5479), while CADE alone contributes a 4.41\% gain (0.5403). When combined, SDIR + CADE achieves the best result (0.5625), yielding an overall 8.70\% improvement over the baseline. These results demonstrate that SDIR and CADE are complementary—SDIR enhances robustness by learning from degraded modality signals, while CADE promotes cross-domain generalization by synthesizing latent target domain distributions using local pathological features and global gene features.

\subsubsection{Synergy analysis}

\begin{table}[]
\caption{Analysis of $\alpha$ in SDIR and $\gamma$ in CADE for synergistic effects. Experiments are conducted on the TCGA benchmark across four cancer types. The model is trained on each single source domain and evaluated on the other target domains using the C-index as the performance metric.}
\label{tab:drop_mix}
\resizebox{0.9\columnwidth}{!}{%
\begin{tabular}{c|c|cccc|c}
\hline \hline
$\alpha$              & $\gamma$              & BLCA                                    & BRCA                                      & HNSC                                    & STAD                                    & Average                                 \\ \hline
0.1                   &                       & 0.5312                                  & 0.5291                                  & \cellcolor[HTML]{E4EEBC}0.5601          & 0.5384                                  & 0.5397                                  \\
0.3                   &                       & 0.5259                                  & \cellcolor[HTML]{BDE6CD}\textbf{0.5613} & 0.5380                                  & \cellcolor[HTML]{FFF8C5}0.5464          & \cellcolor[HTML]{FFF8C5}0.5429          \\
0.5                   &                       & \cellcolor[HTML]{BDE6CD}\textbf{0.5546} & \cellcolor[HTML]{FFF8C5}0.5354          & \cellcolor[HTML]{BDE6CD}\textbf{0.5714} & \cellcolor[HTML]{E4EEBC}0.5506          & \cellcolor[HTML]{BDE6CD}\textbf{0.5530} \\
0.7                   &                       & \cellcolor[HTML]{FFF8C5}0.5486          & 0.5069                                  & \cellcolor[HTML]{FFF8C5}0.5474          & 0.5237                                  & 0.5317                                  \\
0.9                   & \multirow{-5}{*}{0.5} & \cellcolor[HTML]{E4EEBC}0.5533          & \cellcolor[HTML]{E4EEBC}0.5496          & 0.5384                                  & \cellcolor[HTML]{BDE6CD}\textbf{0.5517} & \cellcolor[HTML]{E4EEBC}0.5483          \\ \hline
                      & 0.1                   & 0.5198                                  & \cellcolor[HTML]{E4EEBC}0.5441          & 0.5646                                  & 0.5190                                  & \cellcolor[HTML]{FFF8C5}0.5369          \\
                      & 0.3                   & \cellcolor[HTML]{BDE6CD}\textbf{0.5648} & \cellcolor[HTML]{BDE6CD}\textbf{0.5576} & \cellcolor[HTML]{BDE6CD}\textbf{0.5762} & \cellcolor[HTML]{BDE6CD}\textbf{0.5514} & \cellcolor[HTML]{BDE6CD}\textbf{0.5625} \\
                      & 0.5                   & \cellcolor[HTML]{E4EEBC}0.5546          & 0.5354                                  & \cellcolor[HTML]{E4EEBC}0.5714          & \cellcolor[HTML]{E4EEBC}0.5506          & \cellcolor[HTML]{E4EEBC}0.5530          \\
                      & 0.7                   & \cellcolor[HTML]{FFF8C5}0.5406          & \cellcolor[HTML]{FFF8C5}0.5398          & 0.5387                                  & 0.5166                                  & 0.5339                                  \\
\multirow{-5}{*}{0.5} & 0.9                   & 0.5355                                  & 0.5085                                  & \cellcolor[HTML]{FFF8C5}0.5694          & \cellcolor[HTML]{FFF8C5}0.5278          & 0.5353                                  \\ \hline \hline
\end{tabular}%
}
\end{table}

To further investigate the interaction between the sparsity parameter $\alpha$ in SDIR and the semantic guidance scalar $\gamma$ in CADE, we conducted an extensive grid search over various combinations of these two parameters. The experimental results are summarized in Table~\ref{tab:drop_mix}. When fixing $\gamma = 0.5$, we observed that the best generalization performance was achieved at $\alpha = 0.5$, with an average C-index of 0.5530. This suggests that moderately enhancing strong features while correcting weak ones can help mitigate the feature degradation problem. On the contrary, performance slightly declined—for instance, when $\alpha = 0.9$, the C-index dropped to 0.5483—indicating a saturation point in the benefit of SDIR. Conversely, by fixing $\alpha = 0.5$, we explored the effect of varying $\gamma$. The best performance was obtained when $\gamma = 0.3$, yielding an average C-index of 0.5625. This result highlights the critical role of fine-grained interpolation in CADE: moderately blending domain statistics enables the generation of plausible target-like distributions. Taken together, these findings reveal a significant synergistic effect between SDIR and CADE.

\subsection{Compatibility with Multimodal Prognosis Methods}
\begin{table}[]
\caption{The compatibility of our proposed method. The impact on generalization performance after incorporating our proposed method into classical multimodal survival prognosis algorithms.  }
\label{tab:compatibility}
\resizebox{0.9\columnwidth}{!}{%
\begin{tabular}{c|cccc|c|c}
\hline \hline
Model/Study & BLCA                                    & BRCA                                      & HNSC                                    & STAD                                    & Average                                 & $\Delta$    \\ \hline
MCAT \cite{mcat}        & 0.4625                                  & \cellcolor[HTML]{BDE6CD}\textbf{0.5375} & 0.4662                                  & 0.5145                                  & 0.4952                                  &                          \\
+ours       & \cellcolor[HTML]{BDE6CD}\textbf{0.5408} & 0.5205                                  & \cellcolor[HTML]{BDE6CD}\textbf{0.5076} & \cellcolor[HTML]{BDE6CD}\textbf{0.5309} & \cellcolor[HTML]{BDE6CD}\textbf{0.5249} & \multirow{-2}{*}{0.0297} \\ \hline
MOTCat \cite{motcat}      & 0.4592                                  & 0.4869                                  & 0.4693                                  & 0.5309                                  & 0.4866                                  &                          \\
+ours       & \cellcolor[HTML]{BDE6CD}\textbf{0.5318} & \cellcolor[HTML]{BDE6CD}\textbf{0.5333} & \cellcolor[HTML]{BDE6CD}\textbf{0.5033} & \cellcolor[HTML]{BDE6CD}\textbf{0.5741} & \cellcolor[HTML]{BDE6CD}\textbf{0.5356} & \multirow{-2}{*}{0.0490} \\ \hline
SURVPATH \cite{survpath}    & 0.5255                                  & 0.5089                                  & 0.5385                                  & 0.4970                                  & 0.5175                                  &                          \\
+ours       & \cellcolor[HTML]{BDE6CD}\textbf{0.5648} & \cellcolor[HTML]{BDE6CD}\textbf{0.5576} & \cellcolor[HTML]{BDE6CD}\textbf{0.5762} & \cellcolor[HTML]{BDE6CD}\textbf{0.5514} & \cellcolor[HTML]{BDE6CD}\textbf{0.5625} & \multirow{-2}{*}{0.0450} \\ \hline \hline
\end{tabular}%
}
\end{table}

To evaluate the generality and compatibility of our proposed method, we integrated it into three representative state-of-the-art multimodal survival prediction frameworks: MCAT, MOTCat, and SURVPATH. The results are presented in Table~\ref{tab:compatibility}. Across all tested cancer types (i.e., BLCA for Bladder Urothelial Carcinoma, BRCA for Breast Invasive Carcinoma, HNSC for Head and Neck Squamous Cell Carcinoma, and STAD for Stomach Adenocarcinoma), our method consistently improved the average C-index, demonstrating enhanced generalization regardless of the underlying multimodal model architecture. Specifically, when applied to MOTCat, our method increased the average C-index from 0.4866 to 0.5356, yielding a 4.90\% gain. Even the already competitive SURVPATH model (C-index of 0.5175) benefited significantly from our integration, reaching a C-index of 0.5625.


\section{Discussion}

To further validate the effectiveness of our proposed modules, SDIR and CADE, in survival analysis, we divided all patients into low-risk and high-risk groups based on the median value of the predicted risk scores. We then performed Kaplan–Meier analysis to visualize the survival events of all patients, as shown in Figure~\ref{fig:KM}. The results demonstrate that low-risk patients predicted by our method actually live longer than high-risk patients, indicating that our approach enables the model to achieve strong generalization and robustness in cross-cancer scenarios.

Despite improvements in prognostic model generalization by SDIR and CADE across cancer types, limitations exist. SDIR's fixed-parameter Bernoulli sparsity mask lacks flexibility for sample-specific modality quality. CADE assumes a Gaussian distribution for latent variables, which may not hold for the complex, multimodal nature of real-world biomedical data, possibly affecting accuracy. Nonetheless, this work advances universally applicable and clinically practical survival prognosis tools, enhancing cross-cancer model generalizability with positive public health implications.


\section{Conclusion}
In this paper, we addressed a practical yet challenging task: Cross-Cancer Single-Domain Generalization for Multimodal Prognosis. Our work primarily focused on two key problems: mitigating the degradation of weak modality features and effectively integrating local pathological image information with global gene expression signals to enhance model generalization. To tackle the first issue, we proposed the Sparse Dirac Information Rebalancer (SDIR), which alleviated weak feature degradation by sparsifying dominant modality features and introducing a Dirac-inspired response function to actively amplify weaker modality signals. To further address domain shifts across cancer types, we introduced the Cancer-Aware Distribution Entanglement (CADE) module. Guided by global gene-level semantics, CADE employed a kernel-smooth statistical fusion path to reorganize local pathological representations in the latent space, forming a structured, cancer-aware distribution. This mechanism improved cross-cancer generalization and provides a promising direction for robust multimodal integration in medical image analysis.

\section{Acknowledgements}
This work was supported by Zhejiang Leading Innovative and Entrepreneur Team Introduction Program (2024R01007), the Natural Science Foundation of Gansu Province, China under Grant 25JRRA666 and in part by the Fundamental
Research Funds for the Central Universities of China under Grant lzujbky-2023-it40, lzujbky-2024-it52.

\bibliographystyle{ACM-Reference-Format}
\balance
\bibliography{sample-sigconf}


\end{document}